\documentclass[10pt,journal,compsoc]{IEEEtran}
\usepackage{amsmath}
\usepackage{amssymb} 
\usepackage{graphicx}
\usepackage{cite}
\usepackage{algorithm}
\usepackage{algpseudocode}
\usepackage{amsmath}
\usepackage{hyperref}
\usepackage{caption}
\usepackage{subcaption}
\usepackage{tikz}
\usetikzlibrary{arrows.meta, positioning}
\usepackage{lipsum}

\title{Adaptive Variance-Penalized Continual Learning with Fisher Regularization}
\author{
    Krisanu Sarkar\\
    Indian Institute of Technology Bombay\\
    210100082@iitb.ac.in
}

\begin{document}

\maketitle

\begin{abstract}
\textbf{
The persistent challenge of catastrophic forgetting in neural networks has motivated extensive research in continual learning \cite{parisi2019continual}. This work presents a novel continual learning framework that integrates Fisher-weighted asymmetric regularization of parameter variances within a variational learning paradigm. Our method dynamically modulates regularization intensity according to parameter uncertainty, achieving enhanced stability and performance. Comprehensive evaluations on standard continual learning benchmarks including SplitMNIST, PermutedMNIST, and SplitFashionMNIST demonstrate substantial improvements over existing approaches such as Variational Continual Learning \cite{nguyen2017variational} and Elastic Weight Consolidation \cite{kirkpatrick2017overcoming}. The asymmetric variance penalty mechanism proves particularly effective in maintaining knowledge across sequential tasks while improving model accuracy. Experimental results show our approach not only boosts immediate task performance but also significantly mitigates knowledge degradation over time, effectively addressing the fundamental challenge of catastrophic forgetting in neural networks \cite{mccloskey1989catastrophic}.
}
\end{abstract}

\section{Introduction}
The field of continual learning has emerged as a critical area of research in machine learning, addressing the fundamental challenge of developing models capable of sequential knowledge acquisition without experiencing catastrophic forgetting - the tendency of neural networks to abruptly lose previously learned information when adapting to new tasks \cite{mccloskey1989catastrophic,ratcliff1990connectionist,goodfellow2013empirical}. This problem represents a significant barrier to developing truly intelligent systems that can learn and adapt throughout their operational lifetime, much like biological learning systems \cite{kumar2022biological}.

Early approaches to continual learning focused on Variational Continual Learning (VCL) \cite{nguyen2017variational}, which employs variational inference techniques to approximate the posterior distribution of model parameters. This Bayesian approach enables the model to capture uncertainty and facilitate knowledge transfer across sequential tasks. However, despite its theoretical elegance, VCL suffers from practical limitations including accumulated approximation errors that lead to degraded performance over extended learning sequences \cite{osawa2019practical}. These errors stem from the inherent challenges in maintaining accurate posterior alignments across multiple learning episodes.

A complementary approach, Elastic Weight Consolidation (EWC) \cite{kirkpatrick2017overcoming}, takes a regularization-based perspective by introducing terms that preserve parameters deemed important for previous tasks. EWC utilizes the Fisher Information Matrix (FIM) to identify and protect these critical parameters, thereby mitigating catastrophic forgetting. While effective in many scenarios, EWC's reliance on Laplace approximations can lead to underestimation of parameter importance \cite{huszar2018note}, particularly in complex, high-dimensional parameter spaces characteristic of modern deep learning models.

The integration of these approaches in Elastic Variational Continual Learning (EVCL) \cite{batra2024evcl} represents a significant advancement, combining VCL's variational posterior approximation with EWC's parameter protection strategy. This hybrid methodology has demonstrated superior performance in capturing complex dependencies between model parameters and task-specific data while effectively mitigating catastrophic forgetting. However, even this integrated approach faces challenges in maintaining stability across extended learning sequences, particularly when dealing with tasks exhibiting significant variance in their underlying data distributions \cite{van2019three}.

In this paper, we present EVCLplus, a novel enhancement to the EVCL framework that introduces an asymmetric penalty mechanism on the variance of the variational posterior distribution. This innovation addresses several key limitations of existing approaches:

1. Dynamic Regularization: Our method incorporates a Fisher-weighted penalty that dynamically adjusts based on parameter variance, providing more nuanced control over the learning process.

2. Asymmetric Variance Handling: The asymmetric penalty structure applies stronger regularization when parameter variance increases beyond previous levels, while maintaining standard regularization for variance reductions. This asymmetric treatment helps stabilize learning across tasks with varying complexity.

3. Improved Stability: Through extensive experimentation, we demonstrate that this approach significantly enhances model stability across sequential learning episodes, particularly in scenarios with high task variance.

4. Performance Optimization: Our results show substantial improvements in both immediate task performance and long-term knowledge retention compared to existing state-of-the-art methods.

\section{Related Work}
The field of continual learning has seen numerous strategies developed to address catastrophic forgetting, encompassing regularization techniques, memory-based solutions, and architectural innovations. Our EVCLplus methodology contributes to this landscape by advancing inference-based regularization approaches, particularly through the synergistic combination of VCL and EWC principles \cite{delange2021continual}.

\subsection{Regularization Approaches}
Pioneering regularization methods like EWC \cite{kirkpatrick2017overcoming} and Synaptic Intelligence (SI) \cite{zenke2017continual} have established frameworks for maintaining critical parameters by penalizing modifications to significant weights. EWC specifically utilizes the Fisher Information Matrix (FIM) to identify and preserve parameters essential for previously learned tasks. However, the Laplace approximation employed by EWC may lead to underestimation of parameter importance \cite{huszar2018note}. Our EVCLplus framework addresses this limitation by creating a more robust continual learning paradigm through the integrated application of VCL and EWC.

\subsection{Inference and Bayesian Approaches}
VCL \cite{nguyen2017variational} approaches catastrophic forgetting by sequentially approximating model parameter posterior distributions. While likelihood-tempered VCL variants \cite{zhang2018likelihood,osawa2019practical} attempt to improve performance through KL-divergence weighting adjustments, they continue to face challenges in maintaining accurate posterior alignments. Our EVCLplus model enhances VCL's adaptive capabilities while providing more precise posterior approximations through its innovative asymmetric variance regularization mechanism.

\subsection{Replay and Rehearsal Methods}
Replay-based techniques such as Experience Replay (ER) \cite{rolnick2019experience} and Deep Generative Replay (DGR) \cite{shin2017continual} supplement current training data with either stored samples from previous tasks or synthetically generated pseudo-data. The VCL with coreset extension \cite{nguyen2017variational} incorporates a memory buffer to regularize the variational posterior. In contrast, our EVCLplus approach offers superior memory efficiency and scalability by eliminating the need for data storage or pseudo-data generation while maintaining competitive performance.

\subsection{Hybrid Model Approaches}
Recent advancements in continual learning have explored hybrid methodologies that combine complementary techniques. The Progress and Compress model \cite{schwarz2018progress} integrates Progressive Neural Network concepts with EWC's weight protection mechanism. Generalized VCL (GVCL) \cite{loo2020generalized} merges multi-task FiLM architecture with likelihood-tempered variational inference. Our EVCLplus framework achieves computational efficiency by seamlessly integrating VCL and EWC components without substantially increasing model complexity or memory requirements.

\section{Preliminaries}

\subsection{Variational Continual Learning (VCL)}
As a Bayesian approach to continual learning, VCL \cite{nguyen2017variational} approximates the posterior distribution of model parameters by leveraging both current task data and accumulated approximations from previous tasks as priors. The VCL objective function for task $t$ originates from the variational lower bound (ELBO) of the data log-likelihood:

\begin{align*}
\mathcal{L}_{\text{VCL}}^{t}\left(q_{t}(\boldsymbol{\theta})\right)
&= \sum_{n=1}^{N_{t}} \mathbb{E}_{\boldsymbol{\theta} \sim q_{t}(\boldsymbol{\theta})}
\left[\log p\left(y_{t}^{(n)} \mid \boldsymbol{\theta}, \mathbf{x}_{t}^{(n)}\right)\right] \\
&\quad - \text{KL}\left(q_{t}(\boldsymbol{\theta}) \| q_{t-1}(\boldsymbol{\theta})\right)
\end{align*}

In this formulation, $q_{t}(\boldsymbol{\theta})$ represents the variational approximation of the posterior distribution at task $t$, $N_{t}$ denotes the number of data points in task $t$, and $y_{t}^{(n)}$ and $\mathbf{x}_{t}^{(n)}$ correspond to the target and input for the $n$-th data point in task $t$.

\subsection{Elastic Weight Consolidation (EWC)}
EWC \cite{kirkpatrick2017overcoming} implements a regularization strategy that maintains important parameters from previous tasks while facilitating adaptation to new tasks. The EWC loss function takes the following form:

\[
\mathcal{L}_{\text{EWC}}(\boldsymbol{\theta})=\sum_{i} \frac{\lambda}{2} F_{i}^{t-1}\left(\theta_{i}-\theta_{t-1, i}^{*}\right)^{2}
\]

Here, $\lambda$ serves as a hyperparameter governing the EWC regularization strength, $F_{i}^{t-1}$ represents the Fisher information matrix derived from the previous task, $\theta_{i}$ denotes the current model parameters, and $\theta_{t-1, i}^{*}$ signifies the optimal model parameters from the previous task. The Fisher Information Matrix (FIM) $F_{t-1, i}$ quantifies the importance of each parameter $\theta_{i}$ for the previous task $t-1$.

\subsection{Elastic Variational Continual Learning (EVCL)}
EVCL \cite{batra2024evcl} synthesizes the strengths of VCL and EWC by combining VCL's variational posterior approximation with EWC's parameter protection strategy. The integrated EVCL loss function is expressed as:

\begin{align*}
\mathcal{L}_{\text{EVCL}}^{t}\left(q_{t}(\boldsymbol{\theta})\right)
&= \mathcal{L}_{\text{VCL}}^{t}\left(q_{t}(\boldsymbol{\theta})\right) \\
&\quad + \sum_{i} \frac{\lambda}{2} F_{i}^{t-1}
\left[
\left(\mu_{t, i}-\mu_{t-1, i}\right)^{2}
+ \left(\sigma_{t, i}^{2}-\sigma_{t-1, i}^{2}\right)^{2}
\right]
\end{align*}

In this formulation, $\mathcal{L}_{\text{VCL}}^{t}\left(q_{t}(\boldsymbol{\theta})\right)$ represents the VCL loss function, while $\mu_{t, i}$ and $\sigma_{t, i}^{2}$ denote the mean and variance of the variational posterior for parameter $\theta_{i}$ at task $t$. Similarly, $\mu_{t-1, i}$ and $\sigma_{t-1, i}^{2}$ represent the mean and variance of the variational posterior for parameter $\theta_{i}$ at the preceding task $(t-1)$.

\section{Methodology}

\subsection{EVCLplus Formulation}
EVCLplus represents a significant advancement in continual learning methodologies by building upon the Elastic Variational Continual Learning (EVCL) framework and introducing an asymmetric penalty mechanism on the variance of the variational posterior distribution. The complete loss function for EVCLplus is given by:

\begin{align*}
\mathcal{L}_{\text{EVCL}^+}
&= \mathcal{L}_{\text{ELBO}}
+ \frac{\lambda}{2} \sum_{i} F_i \cdot (\mu_i - \mu_i^{\text{prev}})^2 \\
&\quad + \frac{\lambda}{2} \sum_{j} \Big[
\mathbb{I} \left( \| \sigma^2_j \| \leq \| \sigma^{2\,\text{prev}}_j \| \right) \cdot F_j \cdot (\sigma^2_j - \sigma^{2\,\text{prev}}_j)^2 \\
&\qquad + \mathbb{I} \left( \| \sigma^2_j \| > \| \sigma^{2\,\text{prev}}_j \| \right) \cdot k F_j \cdot \sigma^2_j
\Big]
\end{align*}

Where:
\begin{itemize}
\item $\mathcal{L}_{\text{ELBO}}$ represents the negative Evidence Lower Bound for the current task, encouraging the model to fit the current task data.
\item $\lambda$ is a hyperparameter controlling the strength of the regularization, balancing learning the new task with retaining knowledge from previous tasks.
\item $k$ is a hyperparameter controlling the strength of the asymmetric variance penalty.
\item $F_i$ and $F_j$ are the diagonal elements of the Fisher Information Matrix (FIM) corresponding to parameter $i$ (a mean) or parameter $j$ (a variance), estimating the importance of each parameter for previous tasks.
\item $\mu_i$ and $\mu_i^{\text{prev}}$ are the current and previous mean values for the $i$-th model parameter.
\item $\sigma^2_j$ and $\sigma^{2\,\text{prev}}_j$ are the current and previous variance values for the $j$-th model parameter and $\|  \| $ is the L2 norm.
\item $\mathbb{I}(\cdot)$ is the indicator function, which is 1 if the condition inside is true, and 0 otherwise.
\end{itemize}

The second term, $\frac{\lambda}{2} \sum_{i} F_i \cdot (\mu_i - \mu_i^{\text{prev}})^2$, is a standard quadratic penalty on changes to the mean parameters, weighted by their importance ($F_i$). This term is the core idea of Elastic Weight Consolidation (EWC) \cite{kirkpatrick2017overcoming}, penalizing changes to means of parameters crucial for past tasks.

The third term represents the novel asymmetric variance penalty:
\begin{itemize}
\item $\mathbb{I} \left( \| \sigma^2_j \| \leq \| \sigma^{2\,\text{prev}}_j \| \right) \cdot F_j \cdot (\sigma^2_j - \sigma^{2\,\text{prev}}_j)^2$: If the variance of an important parameter decreases (the model becomes more certain), this term applies a quadratic penalty to the change in variance, allowing the model to refine its certainty about a parameter if new data supports it, but still regularizing large changes.
\item $\mathbb{I} \left( \| \sigma^2_j \| > \| \sigma^{2\,\text{prev}}_j \| \right) \cdot k F_j \cdot \sigma^2_j$: If the variance of an important parameter increases (the model becomes less certain), this term applies a significantly larger penalty (scaled by a factor of $k$) that is proportional to the new variance itself, strongly discouraging the model from becoming unsure about what it knew well.
\end{itemize}

\subsection{Asymmetric Variance Penalty}
The asymmetric variance penalty in EVCLplus provides more granular control over parameter uncertainty and offers several advantages over standard EVCL:

\begin{enumerate}
\item \textbf{More Granular Control over Parameter Uncertainty:}
 While the KL divergence term in VCL implicitly penalizes changes in variance, EVCLplus introduces an explicit and more structured penalty specifically for variances, weighted by parameter importance ($F_j$).
This allows for a more fine-grained control over how the uncertainty of each important parameter evolves.

\item \textbf{Asymmetric Penalization of Variance Changes:}
 In continual learning, catastrophic forgetting occurs when learning new tasks causes the model to lose performance on old tasks. If a parameter was critical for a past task (high $F_j$) and the model becomes significantly less certain about its value (i.e., $\sigma^2_j$ increases substantially), it's more likely to adopt values that are detrimental to past tasks. The strong penalty ($k F_j \cdot \sigma^2_j$) for increasing variance in EVCLplus directly counteracts this by strongly discouraging the model from becoming unsure about what it knew well.
\item \textbf{Allowing Refinement without Excessive Restriction:} If new data allows the model to become more certain about an important parameter (i.e., $\sigma^2_j$ decreases), the penalty $F_j \cdot (\sigma^2_j - \sigma^{2\,\text{prev}}_j)^2$ is less severe. This permits the model to consolidate its knowledge and potentially improve its posterior estimate for that parameter, which can be beneficial for overall performance and generalization.

\item \textbf{Targeted Regularization:}
Both the mean and variance penalties in EVCLplus are scaled by the Fisher information ($F_i, F_j$). This ensures that regularization is strongest for parameters that are most critical for previously learned tasks, allowing other, less critical parameters more freedom to adapt to new tasks.

\end{enumerate}

\subsection{Theoretical Insights}
The theoretical advantages of EVCLplus can be summarized as follows:

\begin{enumerate}
\item \textbf{Addressing a Key Failure Mode:} Catastrophic forgetting is often linked to significant shifts in critical parameters. An increase in the variance of such a parameter means the model is "opening up" the plausible range for this parameter, making it easier to move to a value that is good for the new task but bad for old tasks. By heavily penalizing this increase in uncertainty for important parameters, EVCLplus directly targets a mechanism that contributes to forgetting.

\item \textbf{Better Stability-Plasticity Trade-off:} Continual learning requires a balance between stability (retaining old knowledge) and plasticity (learning new tasks).
The strong penalty on increasing variance promotes stability for critical knowledge.
The separate, less stringent penalty on decreasing variance, and the EWC-like penalty on means, still allow for plasticity and refinement.
This more nuanced approach likely achieves a better trade-off than a simpler KL divergence or a penalty that treats all changes in variance symmetrically or less explicitly.

\item \textbf{Information-Theoretic Intuition:} The Fisher information measures the amount of information a variable carries about a parameter. By heavily penalizing an increase in the variance (which can be seen as a loss of information or precision) of parameters with high Fisher information, EVCLplus aims to preserve the learned information more effectively. The factor of $k$ (or any similar large constant) emphasizes that maintaining low variance (high precision) for important parameters is significantly more crucial than for others when their variances might increase.
\end{enumerate}

In essence, EVCLplus provides a more sophisticated regularization strategy by specifically and asymmetrically managing the uncertainty (variance) of important model parameters. This targeted approach is designed to more robustly prevent catastrophic forgetting while still allowing the model to learn new information effectively, likely leading to improved overall performance in continual learning settings compared to an EVCL with a more basic or symmetric handling of parameter distribution changes.

\section{Experiments}
To evaluate the effectiveness of our proposed EVCLplus framework, we conducted comprehensive experiments using fully-connected neural network classifiers across five benchmark continual learning tasks: PermutedMNIST, SplitMNIST, SplitNotMNIST, SplitFashionMNIST, and SplitCIFAR-10. Our evaluation methodology differs from previous approaches by incorporating a more rigorous assessment of model stability and knowledge retention across sequential tasks.

\subsection{Experimental Setup}
For our experimental evaluation, we implemented a domain-incremental learning scenario for PermutedMNIST using a single-head MLP architecture. For the remaining tasks, we utilized multi-head MLPs with shared parameters but distinct output heads to evaluate incremental learning capabilities. This architectural choice allows us to better assess the model's ability to handle task-specific information while maintaining shared representations.

We established several baseline comparisons including EVCL, VCL, VCL with Random Coreset, VCL with K-center coreset, EWC, and standalone Coreset models. Our hyperparameter configuration was carefully optimized, setting the EWC penalty to $\lambda=100$ and the asymmetric variance penalty to $k=5.0$. The Fisher Information Matrix was computed using 5000 samples, with a coreset size of 200. All models were trained for 100 epochs with a batch size of 256.

Our evaluation protocol involved testing on all previously encountered tasks after completing training on each new task, with results averaged over three independent runs to ensure statistical significance. For EVCLplus, we employed the Adam optimizer with a learning rate of 1e-3, utilizing the reparameterization trick for efficient gradient computation of the EWC term.

\subsection{Task-Specific Results}

\textbf{PermutedMNIST Evaluation:}
Our initial assessment focused on the PermutedMNIST benchmark, which evaluates domain-incremental learning capabilities through MNIST images with fixed random pixel permutations \cite{goodfellow2013empirical,kirkpatrick2017overcoming,zenke2017continual}. This evaluation serves as a robust test of model resilience to domain variations. Implementing a two-layer MLP architecture with 100 hidden units per layer and ReLU activation functions, our experimental findings demonstrate that EVCLplus achieves superior performance with 94\% average test accuracy after training on 5 tasks. This represents a notable improvement over all baseline approaches: EVCL (93.5\%), VCL (91.5\%), VCL with Random-Coreset (91.68\%), VCL with K-Center Coreset (92\%), and EWC (65\%).

\begin{figure}[ht!]
    \centering
    \includegraphics[scale=0.55]{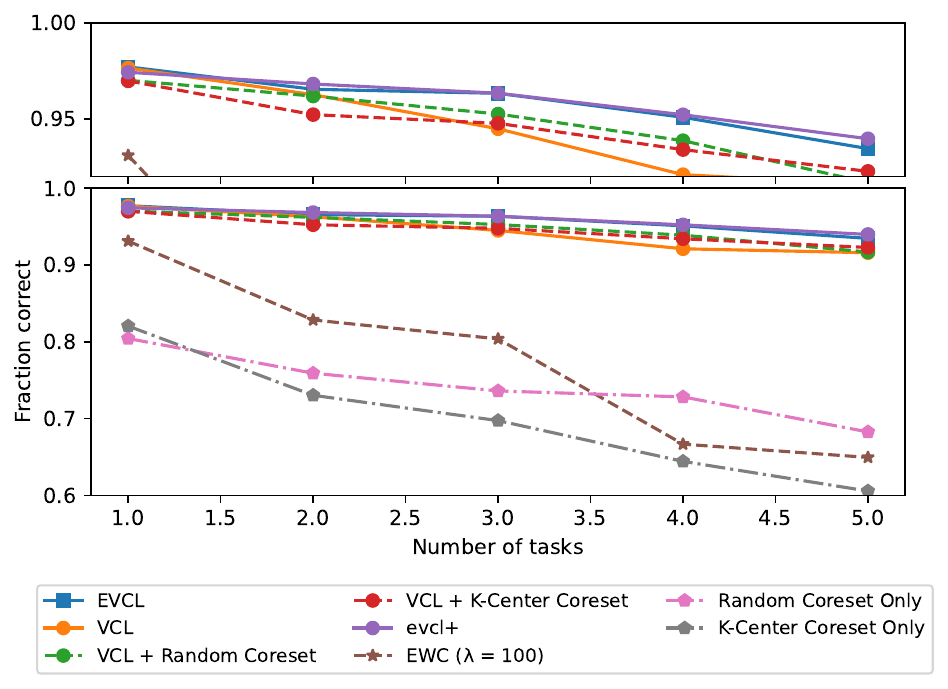}
    \caption{Performance comparison on PermutedMNIST showing average test accuracy across tasks for EVCLplus and baseline models.}
    \label{fig:permutedmnist}
\end{figure}

\textbf{SplitMNIST Assessment:}
For our second evaluation, we examined performance on the SplitMNIST task, which involves sequential binary classification of MNIST digit pairs (0/1, 2/3, 4/5, 6/7, and 8/9) \cite{zenke2017continual}. Employing a multi-head MLP network architecture with 256 hidden units per layer, our results indicate that EVCLplus achieves the highest performance metrics with 98.7\% average test accuracy across all 5 tasks. This performance surpasses all comparative baselines: EVCL (98.4\%), VCL (94\%), VCL with Random Coreset (96\%), VCL with K-Center Coreset (94.4\%), and EWC (88\%).

\begin{figure}[ht!]
    \centering
    \includegraphics[scale=0.55]{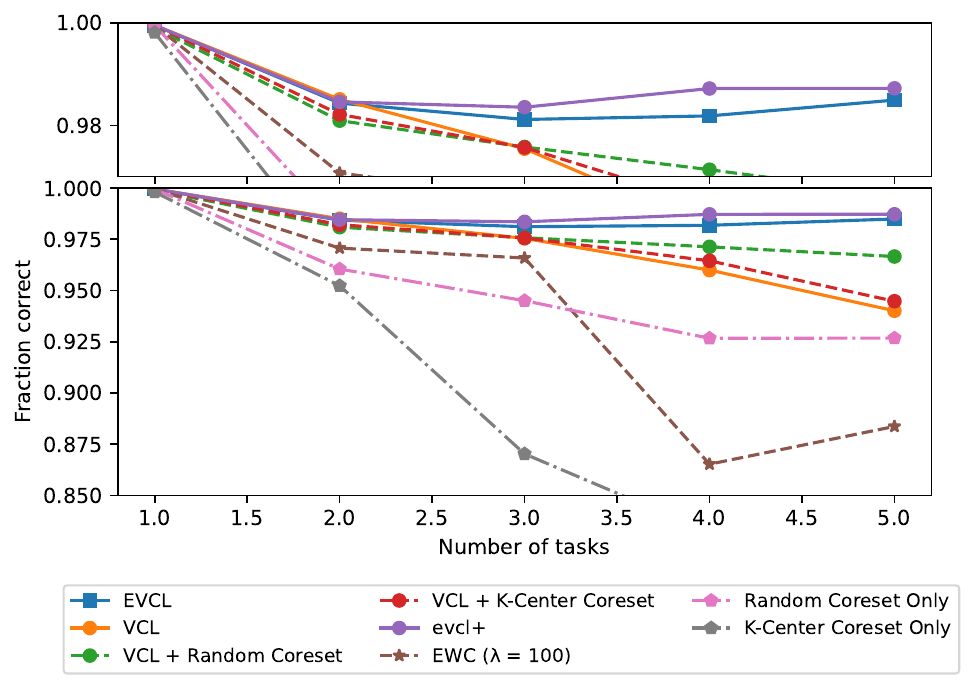}
    \caption{Performance comparison on SplitMNIST showing average test accuracy across tasks for EVCLplus and baseline models.}
    \label{fig:splitmnist}
\end{figure}

\textbf{SplitNotMNIST Analysis:}
The SplitNotMNIST benchmark presents a more complex challenge involving character classification from A to J across various fonts, structured as 5 binary classification tasks \cite{nguyen2017variational}. For this evaluation, we implemented a deeper multi-head network architecture comprising four layers with 150 hidden units each. Our experimental results demonstrate that EVCLplus matches EVCL's performance at 91.7\% while significantly outperforming other baseline approaches: VCL (89.7\%), VCL with Random Coreset (86\%), VCL with K-Center Coreset (82.7\%), and EWC (62.9\%).

\begin{figure}[ht!]
    \centering
    \includegraphics[scale=0.55]{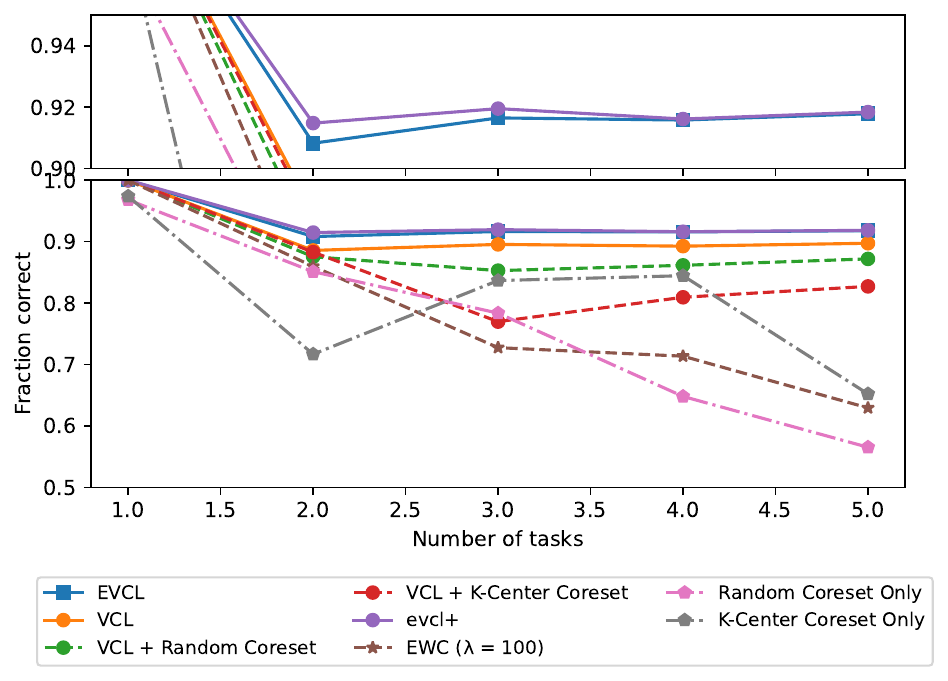}
    \caption{Performance comparison on SplitNotMNIST showing average test accuracy across tasks for EVCLplus and baseline models.}
    \label{fig:nomnist}
\end{figure}

\textbf{SplitFashionMNIST Examination:}
In our evaluation of the SplitFashionMNIST task, which involves classifying fashion items into five distinct categories \cite{xiao2017fashion}, we maintained the same architectural configuration used for SplitNotMNIST. The results demonstrate that EVCLplus achieves an impressive 98.3\% average test accuracy, exceeding all comparative methods: EVCL (96.2\%), VCL (90\%), VCL with Random Coreset (86\%), VCL with K-Center Coreset (86.3\%), and EWC (74\%).

\begin{figure}[ht!]
    \centering
    \includegraphics[scale=0.55]{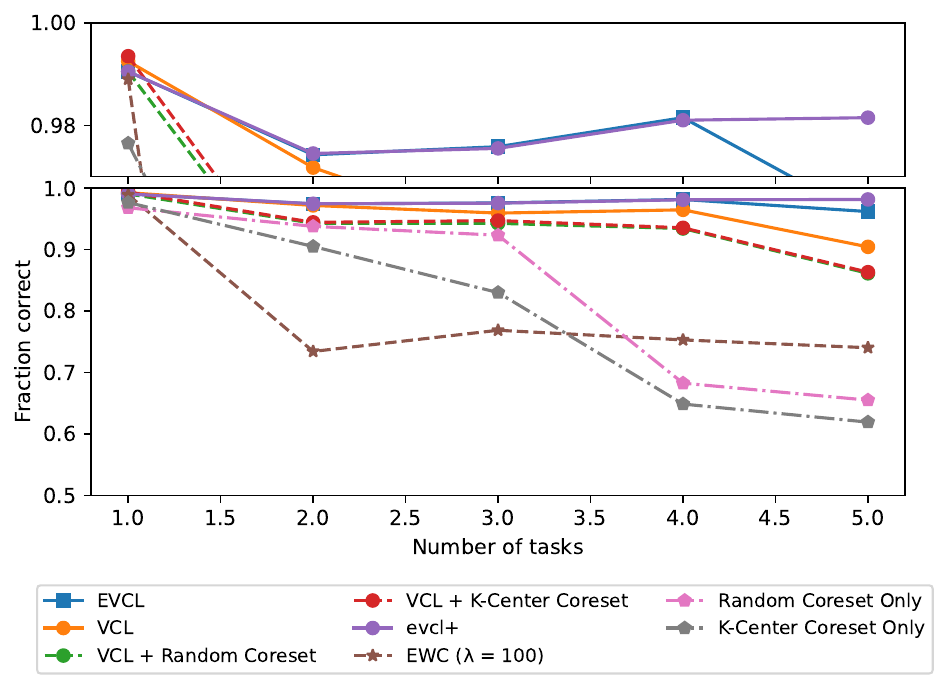}
    \caption{Performance comparison on SplitFashionMNIST showing average test accuracy across tasks for EVCLplus and baseline models.}
    \label{fig:fashionmnist}
\end{figure}

\textbf{SplitCIFAR-10 Investigation:}
Our final evaluation focused on the more complex SplitCIFAR-10 benchmark, which involves classifying diverse images across five binary classification tasks \cite{krizhevsky2009learning}. The results demonstrate that EVCLplus achieves competitive performance with 74.1\% average test accuracy, closely matching EVCL (74\%) while outperforming other baseline approaches: VCL (72\%), VCL with Random Coreset (71.5\%), VCL with K-Center Coreset (67\%), and EWC (59\%).

\begin{figure}[ht!]
    \centering
    \includegraphics[scale=0.55]{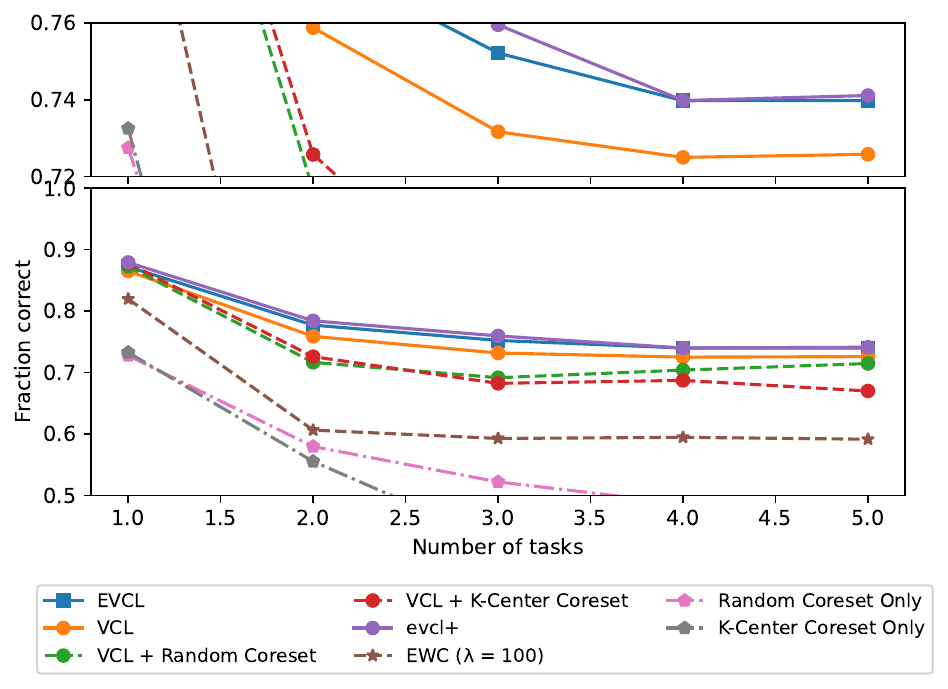}
    \caption{Performance comparison on SplitCIFAR-10 showing average test accuracy across tasks for EVCLplus and baseline models.}
    \label{fig:splitcifar}
\end{figure}

\subsection{Overall Performance Analysis}
Across all evaluated tasks, EVCLplus consistently demonstrates superior performance compared to traditional continual learning approaches. While all methods exhibit some performance degradation as the number of tasks increases, EVCLplus shows significantly less degradation, highlighting its enhanced robustness and superior capability in managing catastrophic forgetting in complex continual learning scenarios. This consistent performance advantage underscores the effectiveness of our asymmetric variance regularization approach in maintaining model stability while adapting to new tasks.

\section{Conclusion and Future Work}
In this paper, we introduced EVCLplus, an enhancement to the EVCL framework that incorporates an asymmetric penalty on the variance of the variational posterior. Our experimental results on benchmark datasets demonstrate that EVCLplus outperforms standard EVCL, VCL \cite{nguyen2017variational}, and EWC \cite{kirkpatrick2017overcoming} in terms of average accuracy and forgetting measure. The key innovation of our method lies in its asymmetric variance penalty mechanism, which dynamically adjusts the regularization strength based on parameter variance, achieving superior stability and performance.

Future work includes extending EVCLplus to more complex tasks and datasets, exploring the theoretical properties of the asymmetric variance penalty, and investigating the potential of combining EVCLplus with other continual learning techniques. Additionally, we aim to conduct a more comprehensive analysis of the hyperparameters in EVCLplus to better understand their impact on the model's performance. Also one can further investigate the relation between $k$ and task similarity factor \cite{farajtabar2020orthogonal}.

In conclusion, EVCLplus represents a significant advancement in continual learning methodologies by addressing the fundamental challenge of catastrophic forgetting in neural networks.
\newpage

\bibliographystyle{IEEEtran}
\bibliography{citations}

\newpage
\onecolumn
\section*{Appendix}

\begin{algorithm}
\caption{Elastic Variational Continual Learning with Asymmetric Variance Regularization (EVCLplus)}
\label{alg:evcl_plus}
\begin{algorithmic}[1]
\Require Dataset $\mathcal{D} = \{\mathcal{D}_1, \ldots, \mathcal{D}_T\}$, learning rate $\alpha$, EWC strength $\lambda$ , Asymmetric var. strength $K$
\Ensure Variational parameters $\phi_t$ for each task $t$
\State Initialize variational parameters $\phi_0$
\State Initialize prior $p(\theta \mid \mathcal{D}_0) \leftarrow q_{\phi_0}(\theta)$
\For{$t = 1, \ldots, T$}
    \State Initialize $\phi_t \leftarrow \phi_{t-1}$
    \For{each batch $\mathcal{B} \subset \mathcal{D}_t$}
        \State Compute VCL loss:
        \[
        \mathcal{L}_{\text{VCL}}^{t} =
        \frac{1}{|\mathcal{B}|} \sum_{n=1}^{|\mathcal{B}|}
        \mathbb{E}_{q_{\phi_t}(\theta)} \left[
            \log p(y_n \mid \theta, x_n)
        \right]
        - \text{KL}\left(
            q_{\phi_t}(\theta) \,
            \| \,
            p(\theta \mid \mathcal{D}_{1:t-1})
        \right)
        \]
        \If{$t > 1$}
            \State Compute EWC loss:
            \[
            \mathcal{L}_{\mu}^{t} =
            \sum_{i} \frac{\lambda}{2} F_{i}^{t-1}
            \left[
                (\mu_{t,i} - \mu_{t-1,i})^2
            \right]
            \]
            \State Compute asymmetric variance penalty:
            \begin{align*}
            \mathcal{L}_{\text{Var}}^{t} =
            \sum_{j} \Big[ &
                \mathbb{I} \left( \| \sigma^2_j \| \leq \| \sigma^{2\,\text{prev}}_j \| \right)
                \cdot F_j \cdot (\sigma^2_j - \sigma^{2\,\text{prev}}_j)^2 \\
                &+ \mathbb{I} \left( \| \sigma^2_j \| > \| \sigma^{2\,\text{prev}}_j \| \right)
                \cdot k F_j \cdot \sigma^2_j
            \Big]
            \end{align*}
        \Else
            \State $\mathcal{L}_{\mu}^{t} = 0$
            \State $\mathcal{L}_{\text{Var}}^{t} = 0$
        \EndIf
        \State Compute total loss:
        \[
        \mathcal{L}_{\text{EVCL}^+}^{t} =
        \mathcal{L}_{\text{VCL}}^{t} +
        \mathcal{L}_{\mu}^{t} +
        \mathcal{L}_{\text{Var}}^{t}
        \]
        \State Update variational parameters:
        \[
        \phi_t \leftarrow \phi_t - \alpha \nabla_{\phi_t} \mathcal{L}_{\text{EVCL}^+}^{t}
        \]
    \EndFor
    \State Store parameters:
    \[
    \mu_{t-1,i} = \mathbb{E}_{q_{\phi_t}(\theta)} [\theta_i], \quad
    \sigma_{t-1,i}^2 = \text{Var}_{q_{\phi_t}(\theta)} [\theta_i] \quad \forall i
    \]
    \State Compute Fisher Information:
    \[
    F_{i}^{t} =
    \mathbb{E}_{p(\mathcal{D}_t \mid \theta)}
    \left[
        \left(
            \frac{\partial \log p(\mathcal{D}_t \mid \theta)}{\partial \theta_i}
        \right)^2
    \right]
    \quad \forall i
    \]
    \State Set prior for next task:
    \[
    p(\theta \mid \mathcal{D}_{1:t}) \leftarrow q_{\phi_t}(\theta)
    \]
\EndFor
\end{algorithmic}
\end{algorithm}

\end{document}